\documentclass{article}

\usepackage{PRIMEarxiv}

\usepackage[utf8]{inputenc} 
\usepackage[T1]{fontenc}    
\usepackage{hyperref}       
\usepackage{url}            
\usepackage{booktabs}       
\usepackage{amsfonts}       
\usepackage{nicefrac}       
\usepackage{microtype}      
\usepackage{lipsum}
\usepackage{fancyhdr}       
\usepackage{graphicx}       
\graphicspath{{media/}}     
\usepackage{amsmath}

\usepackage{algorithm}
\usepackage{algpseudocode}
\usepackage{amsfonts}
\floatname{algorithm}{Algorithm}
\usepackage{setspace} 

\title{Graph Convolutional Network for Semi-Supervised Node Classification With Subgraph Sketch}

\author{
  Zibin Huang\\
  School of Systems Science and Engineering \\
  Sun Yat-sen University \\
  Guangzhou\\
  \texttt{huangzb5@mail2.sysu.edu.cn} \\
   \And
  Jun Xian  \\
  School of Mathematics  \\
  Sun Yat-sen University \\
  Guangzhou\\
  \texttt{xianjun@mail.sysu.edu.cn} \\
}

\begin{document}
\maketitle

\begin{abstract}
   In this paper, we propose the \underline{G}raph-\underline{L}earning-\underline{D}ual \underline{G}raph \underline{C}onvolutional Neural \underline{N}etwork called GLDGCN based on the classical message passing Graph Convolutional Neural Network(GCN) by introducing dual convolutional layer and graph learning layer. We apply GLDGCN to the semi-supervised node classification task. Compared with the baseline methods, we achieve higher classification accuracy on three citation networks Citeseer, Cora and Pubmed, and we also analyze and discuss about selection of the  hyperparameters and network depth. GLDGCN also perform well on the classic social network KarateClub and the new Wiki-CS dataset.

   For the insufficient ability of our algorithm to process large graph data during the experiment, we also introduce subgraph clustering and stochastic gradient descent technology into GCN and design a semi-supervised node classification algorithm based on the \underline{CL}ustering \underline{G}raph \underline{C}onvolutional neural \underline{N}etwork, which enables GCN to process large graph and improves its application value. We complete semi-supervised node classification experiments on two classical large graph which are PPI (more than 50,000 nodes) and Reddit (more than 200,000 nodes), and also perform well.
\end{abstract}

\section{INTRODUCTION}
   Graphs are network structures that are ubiquitous in the real world and used to represent connected data\cite{singh2014application}, such as social networks, gene regulatory networks, ecosystems, knowledge maps, and information system, recommendation system, etc. With the development of artificial intelligence technology and the increasing needs of life, graph-based machine learning has received increasing attention from researchers\cite{isufi2024graph}. There are many applications related to graph data, including graph classification, link prediction, graph recognition, distributed computing and so on.

   In recent years, graph neural network(GNNs) has emerged as a major tool for graph machine learning\cite{defferrard2016convolutional}\cite{gilmer2017neural}\cite{velivckovic2017graph}\cite{xu2018graph}\cite{manessi2020dynamic}, which has found numerous applications in scenarios such as computer vision\cite{shi2020point}, protein design\cite{Tunyasuvunakool2021HighlyAP}, finance\cite{xiang2022temporal},drug discovery\cite{zhang2023emerging} and so on. Node classification task is also one of the most important applications in the field of graph signal processing(GSP) and graph convolutional nerual network(GCN)\cite{sandryhaila2014discrete}\cite{gama2020graphs}, in which GCN is used to implement semi-supervised node classification to achieve good effect and reduce the cost of labeling\cite{kipf2016semi}\cite{atwood2016diffusion}\cite{zhuang2018dual}\cite{zhu2020gssnn}. The structure and generalization ability of GCN are the points that researchers care about. Additionally, training deeper GCN, especially for large graphs are also of great significance\cite{hamilton2017inductive}\cite{huang2018adaptive}\cite{chiang2019cluster}\cite{zeng2019graphsaint}\cite{bai2021ripple}. Despite the success, the improvement of its generalizaion ability\cite{monti2017geometric} and scaling up GCNs to large graphs remain\cite{zeng2021decoupling}\cite{huang2021scaling}the main challenges, which limit the usage.

   In this paper, we propose the Graph Learning Dual Graph Convolutional Neural Network called GLDGCN based on the classical Message Passing Graph Convolutional Neural Network by introducing dual convolutional layer and graph learning layer and apply it to the semi-supervised node classification task. GLDGCN extracts the PPMI matrix of graph through the dual convolution layer, which is a supplement to the feature of adjacency matrix. As for the graph learning layer, it helps GCN accept the input of general matrix data, which extend its application. In order to solve the problem that the space computation complexity is too high when GLDGCN processes large graphs, we also introduces the subgraph clustering technology and stochastic gradient descent technology into GCN, and designs a large-graph-oriented cluster GCN which has the ability to process large data, and the advantages and disadvantages of mini-batch processing are also discussed. Our contributions are summarized as follows:

   (1) A graph convolutional neural network GLDGCN is designed. The input graph can be enhanced to generate a better graph structure through the graph learning layer. For general matrix form data, the graph learning layer can also generate a reasonable graph structure for it as an input to the convolution layer. The design of dual convolutional layer helps  GCN extract features more comprehensively, and enhances the ability in semi-supervised node classification. GLDGCN has achieved higher classification accuracy compared to the baseline method on Citeseer, Cora, Pubmed, and the classic KarateClub social network and the newer Wiki-CS network.

   (2) For the lack of GLDGCN to process large graphs, we introduce subgraph clustering and stochastic gradient descent technology into GCN, which makes GCN have the ability to process large graphs through mini-batch training. The algorithm finishes the semi-supervised node classification task well on the Pubmed dataset (19719 nodes) and the classic large graphs PPI dataset (> 50,000 nodes) and Reddit dataset (> 200,000 nodes).And the advantages and disadvantages of mini-batch processing and some optimization directions are also proposed.

\section{PRELIMINARY}
\subsection{Graph and Matrix Notations}

In this paper, all graphs considered are undirected. A graph is $\mathcal{G} = (\mathcal{V}, \mathcal{E})$ with  $N$ nodes $\in \mathcal{V}$, and $M$ edges $(i, j) \in \mathcal{E}$. The adjacency matrix is denoted as $\bf{A}$ = $[a_{ij}] \in \{0,1\}^{N \times N}$, which satisfies $a_{ij} = a_{ji}$, and the degree matrix is $\bf{D}$ with $D_{i,i} = \sum_{j} A_{i,j}$. The laplacian matrix of $\mathcal{G}$ is denoted as $\bf{L}$ = $I_N-D^{-\frac{1}{2}}AD^{-\frac{1}{2}}$.

A graph signal $x$ is a function that maps elements from a set of points to real numbers, i.e. $x$: $\mathcal{V} \rightarrow \mathbb{R}$. $x \in \mathbb{R}^N$ and $x_i$ is the value of signal in point $i$. The eigendecomposition of laplacian matrix is defined as  $\bf{L}$ = $I_N - D^{-1/2}AD^{-1/2} = U \Lambda U^T$. $\Lambda$ is diagonal matrix, remarked by $\Lambda = diag(\pmb{\lambda})$, and $\pmb{\lambda} = [\lambda_1,\cdots,\lambda_N]$.

\subsection{Graph Fourier Transform and the Convolution on Graph}
The fourier transform of a graph signal $x$ is denoted as
\begin{equation}
    \Hat{x} = \mathcal{F}(x) = U^Tx.
\end{equation}
Correspondingly, the inverse fourier transform is defined as $x = U\Hat{x}$ because $U$ is orthogonal matrix\cite{shuman2013emerging}.

Supposing $y \in \mathbb{R}^N$ and $x \in \mathbb{R}^N$ are graph signals. Based on the Convolution Theorem\cite{deitmar2014principles}, the convolution on graph is denoted as  
\begin{equation}
    y * x = \mathcal{F}^{-1}(\mathcal{F}(y) \cdot \mathcal{F}(x).
\end{equation}
Let $g_{\theta} = \mathcal{F}(y) = diag(\theta)$. $\theta \in \mathbb{R}^N$ are learnable parameters. the spectral convolution on graph can be denoted as 
\begin{equation}
    g_{\theta} * x  = \mathcal{F}^{-1}(g_{\theta} \cdot \mathcal{F}(x)) = Ug_{\theta}U^Tx.
\end{equation}
$U$ is composed of the eigenvectors of the laplacian matrix $\bf{L}$, i.e. $\bf{L}$ = $I_N - D^{-1/2}AD^{-1/2}$ = $U \Lambda U^T$. Therefore, $g_{\theta}$ can be regarded as a function about the eigenvalue of $\bf{L}$, remarked as $g_{\theta}(\Lambda)$.

\subsection{Semi-supervised Node Classification Based on GCN}
It can be seen that in the process of calculating the convolution, $\bf{U}$ composed of the eigenvectors of $\bf{L}$ is needed, so the eigendecomposition must be carried out first. Once there is any change in the graph data, it needs to be recalculated.

The time complexity of calculating $\bf{L}$ is $\mathcal{O}(N^3)$. This is because the LU decomposition \cite{bartels1969simplex} need to carry out a series of operations on the original matrix to decompose it as the product of the lower triangular matrix (L) and the upper triangular matrix (U). This process is usually done using Gaussian elimination. At each step, we need to linearly combine one row of the matrix to eliminate specific elements in the other rows, thereby gradually converting the matrix into an upper triangular form. For a matrix of $N \times N$, we need to iterate $N$ times. In each iteration, we need to perform $2N^2$ operations to update the elements.

Therefore, the time complexity of the decomposition process is $\mathcal{O}(N^3)$, which is unfriendly to large-scale graphs.

To overcome the above problems, $g_\theta(\Lambda)$ can be approximated by truncating polynomial expansion, for example, using a monomial basis, or as proposed by Hammond et al. (2011)\cite{hammond2011wavelets}, Using the Chebyshev polynomial $T_k(x)$ of order $k$.

Let $\widetilde{\Lambda}$ = $\frac{2}{\lambda_{max}}\Lambda$ - $\bf{I}_N$. $\lambda_{max}$ is the maximum eigenvalue of $\bf{L}$. $\theta' \in \mathbb{R}^K$ are polynomial coefficients. $T_0(x) = 1$, $T_1(x) = x$, we have
\begin{equation}
    g_{\theta'}(\Lambda) \approx \sum_{k=0}^K \theta_k^{'}T_k(\widetilde{\Lambda}).
\end{equation}
and

\begin{equation}
    T_k(x) = 2xT_{k-1}(x) - T_{k-2}(x).
\end{equation}
According to (3) and (4), we now have
\begin{equation}
    g_{\theta^{’}} * x  \approx \sum_{k=0}^K \theta_k^{'}T_k(\widetilde{\Lambda})x.
\end{equation}
In (6), $\widetilde{L}$ = $\frac{2}{\lambda_{max}}L$ - $I_N$, and $(U \Lambda U^T)^k$ = $U {\Lambda}^k U^T$ because $U$ is orthogonal matrix. The convolution is $K$- localized because it is a $K$- order polynomial approximation of $g_{\theta}(\Lambda)$. That is to say, it only depends on the nodes that is not more than $K$- hop from the central node, so it can be considered a spatial filter. The computational complexity of (6) is $\mathcal{O}(K|\mathcal{E}|)$, which is a linear computational complexity algorithm about the number of edges. Defferrard use this $K$- localized convolutional filter to design a graph convolution neural network Chebynet\cite{defferrard2016convolutional}.

Let $h_i^{(l)} \in \mathbb{R}^{d_l}$ be the representation vector of the $l$-th hidden layer of node $i \in \mathcal{V}$ with the dimension of $d_l$. In GCN, the forward propagation can be expressed in the following form
\begin{equation}
    h_i^{(l+1)} = \sigma(W_0^{(l)^T}h_i^{(l)} + \sum_{j \in \mathcal{N}_i} c_{i,j}{W}_1^{(l)^T}h_j^{(l)}).
\end{equation}
Where $\sigma$ is a nonlinear activation function, such as the ReLU activation function. $W_0^{(l)}$ and $W_1^{(l)}$ are learnable parameters matrix, $\mathcal{N}_i$ is the set of the First-order neighbor of node $i$, $c_{i,j} = 1/ \sqrt{D_{i,i}D_{j,j}}$ are the normalized constants, $D_{i,i}$ is the degree of the node $i$.

Let $c_{i,j} = D^{-\frac{1}{2}} A D^{-\frac{1}{2}}$, $H^{(l)} \in \mathbb{R}^{N \times d_l}$ is the output matrix of $H^{(0)} = X$ after $l$-th layer, the forward propagation steps in message passing GCN in matrix form is 
\begin{equation}
    H^{(l+1)} = \sigma(H^{(l)}W_0^{(l)} + D^{-\frac{1}{2}} A D^{-\frac{1}{2}} H^{(l)} W_1^{(l)}).
\end{equation}
At this point, we can clearly see the connection between the GCN message passing step and the definition of the approximate spectral convolution in (6): Let $K = 1$, taking a first order approximation of the spectral convolution, and then approximate $\lambda_{max} \approx 2$,and we have 
\begin{equation}
    g_{\theta ^{'}} \star x \approx \theta^{'}_0 x + \theta^{'}_1(L-I_N)x = \theta^{'}_0 - \theta^{'}_1 D^{-\frac{1}{2}} A D^{-\frac{1}{2}}x.
\end{equation}
In semi-supervised learning, it is easy to overfit labeled nodes because the training samples used for learning are small. This is generally solved by using a single parameter matrix $W^{(l)}$\cite{kipf2016semi}.
\begin{equation}
    H^{(l+1)} = \sigma((I_N + D^{-\frac{1}{2}} A D^{-\frac{1}{2}}) H^{(l)} W_1^{(l)}).
\end{equation}
The eigenvalue of the operator $I_N + D^{-\frac{1}{2}} A D^{-\frac{1}{2}}$ is in the range of $[0,2]$. It is found that the stability of the deep neural network will be affected when the operator is repeatedly applied to train the model. Normalizing the adjacency matrix as $I_N + D^{-\frac{1}{2}} A D^{-\frac{1}{2}}$ = $\widetilde{D}^{-\frac{1}{2}} \widetilde{A} \widetilde{D}^{-\frac{1}{2}}$, where $\widetilde{A} = A + I_N$, $\widetilde{D}_{i,i} = \sum_j \widetilde{A}_{i,j}$ can have a positive impact on network performance. The single-parameter message passing in GCN then becomes:
\begin{equation}
    H^{(l+1)} = \sigma(\widetilde{D}^{-\frac{1}{2}} \widetilde{A} \widetilde{D}^{-\frac{1}{2}}H^{(l)} W_1^{(l)}).
\end{equation}
Firstly, we consider using two-layer GCN for semi-supervised node classification on a graph with symmetric adjacency matrix $\bf{A}$. The forward propagation model adopts the following simple form
\begin{equation}
   Z = f(X,A) = softmax(\Hat{A}ReLU(\Hat{A}XW^{(0)})W^{(1)}),
\end{equation}

\begin{figure}[ht]
\centering
\includegraphics[width=0.5\textwidth]{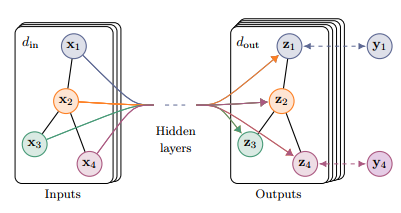}
\caption{The forward propagation model of GCN\cite{kipf2016semi}} 
\end{figure}
$W^{(0)} \in \mathbb{R}^{d_{in} \times d_{hid}}$ is parameter matrix with $H$ input-hidden layer feature mappings. And $W^{(1)} \in \mathbb{R}^{d_{hid} \times d_{out}}$ is parameter matrix of hidden-output layer. The softmax function is denoted by row $softmax(x_i) = \frac{1}{\mathcal{Z}}exp(x_i)$. $\mathcal{Z} = \sum_i exp(x_i)$. As Figure 1 shown, $x_i$ denotes the input data of GCN, $z_i$ denotes the output, $y_i$ denotes the labeled samples.

We use cross-entropy losses on all labeled nodes to optimize the semi-supervised node classification task on GCN
\begin{equation}
\mathcal{L}_{0} = -\sum_{l \in \mathcal{Y}_L} \sum_{f=1}^{d_{out}} \mathcal{Y}_{l,f}ln Z_{l,f}.
\end{equation}
Where $\mathcal{Y}_L$ is the set of labeled node indicators, and $Y_{l,f}$ is an indicator variable that is 1 if the node $l$ is labeled, and 0 otherwise. The weights matrix $W(0)$ and $W(1)$ were trained by the whole-batch gradient descent. During each training parameter update, gradient descent is performed using the complete data set. Similar to the spectral convolution method, the time complexity of GCN is $\mathcal{O}(|\mathcal{E}|)$, which is the first-order linear complexity of the number of edges.

\subsection{Dual Graph Convolutional Neural Network}
The features of the graph can be recorded not only by the adjacency matrix $A \in \mathbb{R}^{n \times n}$, but also by combining the idea of random walk. Define the PPMI (Positive Pointwise Mutual Information) matrix of the graph $P \in \mathbb{R}^{n \times n}$, before calculating the PPMI matrix of the graph, We need to compute the frequency matrix of the graph $F \in \mathbb{R}^{n \times n}$.

The Markov chain that describes a sequence of nodes of random walk is called a random walk. If the random walk is on the node $x_i$ at time $t$, the state is defined as $s(t) = x_i$. From the current node $x_i $ to adjacent nodes, the transition probability can be remark as $p(s(t + 1) = x_j |s(t) = x_i)$, and we have
\begin{equation}
p(s(t + 1) = x_j |s(t) = x_i) = \frac{A_{i,j}}{\sum_j A_{i,j}}.
\end{equation}
We compute the frequency matrix $F \in \mathbb{R}^{n \times n}$ of the graph by the following algorithm
\begin{algorithm}[ht]
\caption{Calculating the frequency matrix of a graph $F \in \mathbb{R}^{n \times n}$ \cite{zhuang2018dual}}
\begin{algorithmic}[1]
    \State \textbf{Input:} Adjacency matrix $A$, parameters: path length $q$, window size $w$, node step length $\gamma$
    \State \textbf{Output:} Frequency matrix $F \in \mathbb{R}^{n \times n}$
    \State initializes the rating matrix $F$ to an all-0 matrix 
    \For{each node $x_i \in \mathcal{V}$} 
        \State Set $x_i$ as the starting point 
        \For{$i = 1$ to $\gamma$} 
            \State $S = \text{RandomWalk}(A,x_i,q)$ 
            \State samples all $(x_n,x_m) \in S$ uniformly with $w$ 
            \For{each $(x_n,x_m)$}
                \State $F_{n,m} = F_{n,m} + 1.0$; $F_{m,n} = F_{m,n} + 1.0$ 
            \EndFor
        \EndFor
    \EndFor
\end{algorithmic}
\end{algorithm}

The time complexity of Algorithm 1 is $\mathcal{O}(n\gamma q^2)$, where $\gamma$ is the node step length and $q$ is the path length, both of which are parameters of Algorithm 1. The $\gamma$ and $q$ are small integers, so $F$ can be computed quickly. We can compute the PPMI matrix of the graph $P \in \mathbb{R}^{n \times n}$ as follows
\begin{equation}
p_{i,j} = \frac{F_{i,j}}{\sum_{i,j}F_{i,j}},
\end{equation}
\begin{equation}
p_{i,*} = \frac{\sum_j F_{i,j}}{\sum_{i,j}F_{i,j}},
\end{equation}
\begin{equation}
p_{*,j} = \frac{\sum_i F_{i,j}}{\sum_{i,j}F_{i,j}},
\end{equation}
\begin{equation}
P_{i,j} = max\{log(\frac{p_{i,j}}{p_{i,*}p_{*,j}}),0\}.
\end{equation}
PPMI has been extensively studied in natural language processing (NLP) \cite{turney2010frequency}\cite{levy2014neural}. In fact, PPMI indicators perform well on semantic similarity tasks \cite{bullinaria2007extracting}. By introducing the PPMI matrix of  graph, the forward propagation layer based on the PPMI matrix in GCN can be further designed as
\begin{equation}
    Z^{(l+1)} = ReLU({D^{(p)}}^{-\frac{1}{2}} P {D^{(p)}}^{-\frac{1}{2}}Z^{(l)} W^{(l)}).
\end{equation}
Where $P$ is the PPMI matrix of graph, $D^{(p)}_{i,i}$ = $\sum_j P_{i,j}$. By using a forward propagation structure similar to GCN, we can design a dual graph convolutional neural network as shown 
\begin{figure}[h]
\centering
\includegraphics[width=0.9\textwidth]{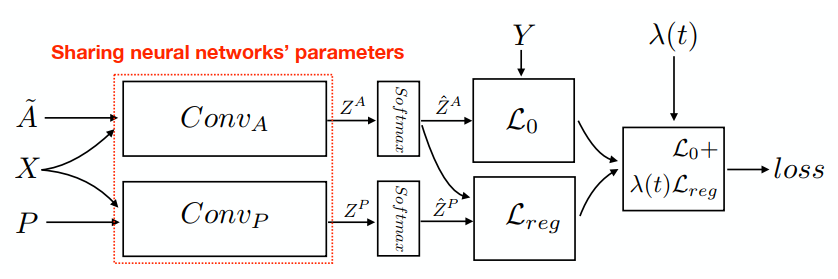}
\caption{Dual Graph Convolutional Neural Network(DGCN)\cite{zhuang2018dual}}
\end{figure}

$\mathcal{L}_0$ is the cross entropy loss function in (13), $\mathcal{L}_{reg}$ is denoted as
\begin{equation}
\mathcal{L}_{reg} = \frac{1}{n}\sum_{f=1}^d||\hat{\mathcal{Z}}^P_{i,:} - \hat{\mathcal{Z}}^A_{i,:}||^2.
\end{equation}
Where $\hat{\mathcal{Z}}^P$and $\hat{\mathcal{Z}}^A$ are respectively the output of the convolutional neural network after the softmax layer, and $\lambda(t)$can be set as the parameter of the neural network, denoted as $\lambda_1$. DGCN needs to minimize the following loss function
\begin{equation}
\mathcal{L}_{DGCN} = \mathcal{L}_{0} + \lambda_1\mathcal{L}_{reg}.
\end{equation}

\section{OUR METHOD}
As you can see from sections 2.3 and 2.4, graph data is represented by $\mathcal{G}(X, A)$ as input to GCN, which is very important. In some applications, the graph structure of the data can be obtained directly in the corresponding domain, such as chemical molecular maps, social networks, etc. In this case, the existing graph structure can be directly input into GCN. However, in other situations, graph structure does not exist, such as image recognition, table data processing, etc. In order to expand the application capability of GCN, a popular method is to artificially build a graph for the data that can be used for GCN input (such as $k$-nearest neighbor graph \cite{favaro2011closed}). However, such artificially estimated graphs are usually independent of the classification process and have high requirement on data quality, so they cannot be guaranteed to serve GCN learning well. In addition, the man-made graph structure is easy to be affected by various factors and is not easy to be unified.

\subsection{Semi-supervised Node Classification Based on Dual Graph Convolutional Neural Network with Graph Learning Layer}
In order to overcome the above problems, the graph learning convolutional neural network can be designed with the graph learning layer and the graph convolutional layer integrated in a unified network architecture, and the adaptive graph representation for GCN input can be learned. As shown in the figure below, a graph learning convolutional neural network consists of a graph learning layer, a dual graph convolutional layer, and an output layer that trains the network parameters by minimizing the loss function.
\begin{figure}[h]
\centering
\includegraphics[width=1.0\textwidth]{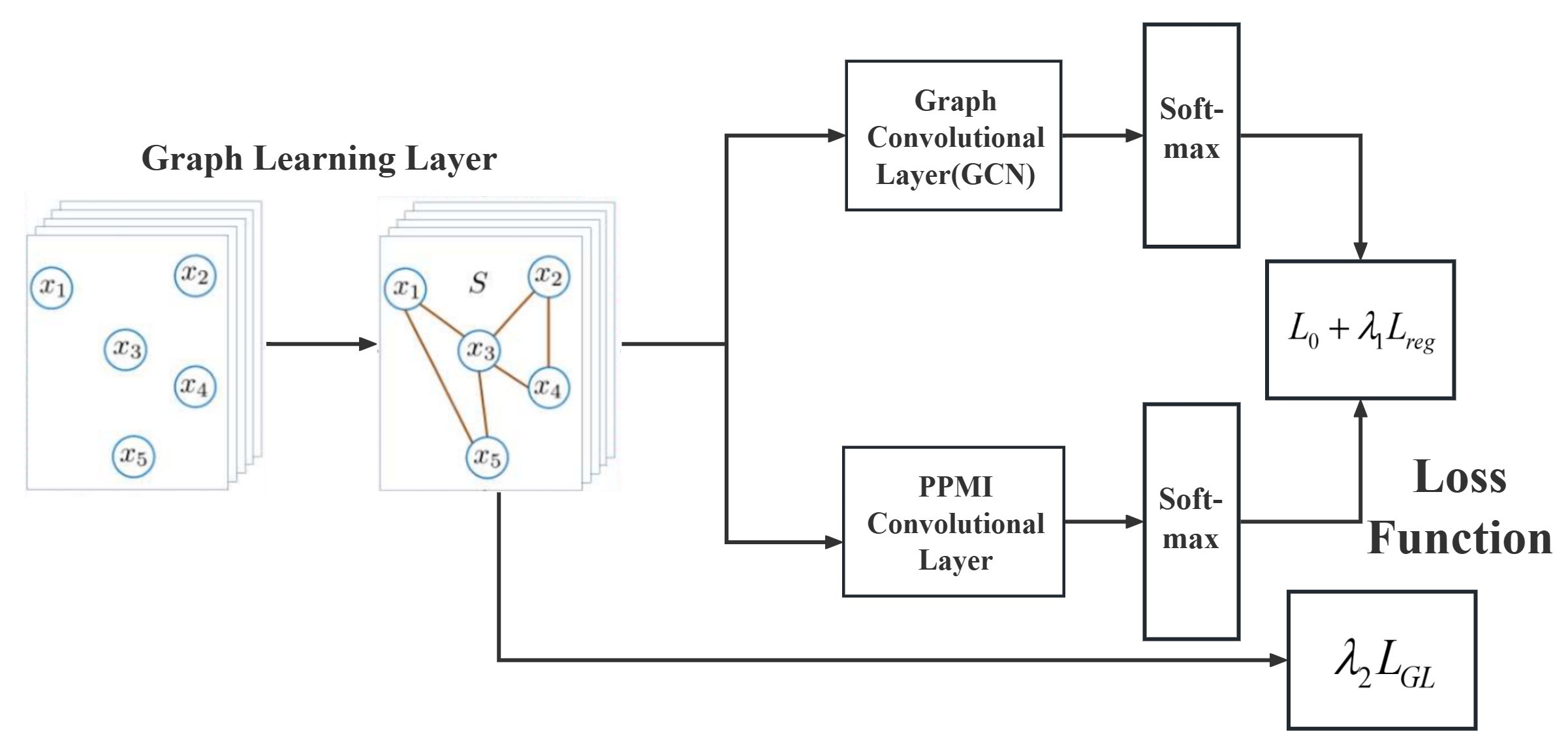}
\caption{Graph Learning Dual Graph Convolutional Neural Network(GLDGCN)}
\end{figure}

The input data $X = (x_1,x_2,...,x_n) \in \mathbb{R}^{n \times p}$. We hope to find a nonnegative function $S_{ij} = h(x_i,x_j)$ to link $x_i$ and $x_j$. Supposing $h(x_i,x_j)$ with parameters $a = (a_1,a_2,...,a_p)^T \in \mathbb{R}^{p \times 1}$, We hope to learn the graph structure $S$ by introducing the Graph Learning Layer in Figure 3.
\begin{equation}
S_{ij} = h(x_i,x_j) = \frac{exp(ReLU(a^T|x_i - x_j|))}{\sum_{j=1}^n exp(ReLU(a^T|x_i - x_j|))}.
\end{equation}
Where $ReLU(x) = max(0,x)$ is an activation function that guarantees that $S_{ij}$ is non-negative. The operation on the right of (22) similar to the softmax function, which guarantees that for any given $i$, $S$ satisfies the column sum of 1, i.e.
\begin{equation}
\sum_{j=1}^nS_{ij} = 1, S_{ij} \geq 0,
\end{equation}
An optimal set of parameters $a = (a_1,a_2,...,a_n)$ can be determined by minimizing the following loss function
\begin{equation}
\mathcal{L}_{GL} = \sum_{i,j=1}^n ||x_i - x_j||^2_2S_{ij} + \gamma||S||^2_F.
\end{equation}
Where $||S||_F = \sqrt{\sum_{i=1}^{n} \sum_{j=1}^{n} |S_{ij}|^2}$ is a regularization term used to control the sparse degree\cite{nie2014clustering}. We can also see from the loss function that large value of $|| x_i-x_j ||_2$ makes smaller value of $S_{ij}$. $\gamma$ is the hyperparameter set in the neural network.

For data that already with a graph structure, that is to say, the adjacency matrix $bf{A}$ exists, then the graph learning layer is defined as
\begin{equation}
S_{ij} = h(x_i,x_j) = \frac{A_{ij}exp(ReLU(a^T|x_i - x_j|))}{\sum_{j=1}^n A_{ij}exp(ReLU(a^T|x_i - x_j|))},
\end{equation}
with the following loss functions need to be minimized.
\begin{equation}
\mathcal{L}_{GL} = \sum_{i,j=1}^n ||x_i - x_j||^2_2S_{ij} + \gamma||S||^2_F + \beta||S-A||^2_F.
\end{equation}
where $\gamma$ and $\beta$ are both hyperparameters that can be set in the neural network.

For the forward propagation layer of GLDGCN, we combined the technique in section 2.3 and 2.4. In Figure 3, the forward propagation rule in the Graph Convolutional Layer(GCN) is as follows
\begin{equation}
    X^{(l+1)} = \sigma({D_s^{(l)}}^{-\frac{1}{2}} S {D_s^{(l)}}^{-\frac{1}{2}}X^{(l)} W^{(l)}).
\end{equation}
$D_s = diag(d_1,d_2,...,d_n)$ is diagonal matrix with diagonal elements $d_i = \sum_{j=1}^n S_{ij}$. $W^{(l)} \in \mathbb {R}^{d_k \times d_{k+1}}$ denotes the weight matrix of $l$-th layer. $\sigma$ denotes the ReLU activation function.

For the PPMI Convolutional Layer, the forward propagation rule is
\begin{equation}
     X^{(l+1)} = \sigma({D_p^{(l)}}^{-\frac{1}{2}} P_s {D_p^{(l)}}^{-\frac{1}{2}}X^{(l)} W^{(l)}).
\end{equation}
$P_s$ denotes the PPMI matrix of $S$, which can be calculated by (15) - (18).

Supposing that $\hat{\mathcal{Z}}^A$ and $\hat{\mathcal{Z}}^P$ are the output of the Graph convolutional neural network(GCN) and PPMI Convolutional layer after softmax layer respectively. We need to minimize the following loss function in Graph Learning Dual Graph Convolutional Neural Networks(GLDGCN)
\begin{equation}
\mathcal{L}_{GLDGCN} = \mathcal{L}_0 + \lambda_1 \mathcal{L}_{reg} + \lambda_2 \mathcal{L}_{GL}.
\end{equation}
In which $\mathcal{L}_0$ denotes the cross entropy loss function shown in (13) and $\mathcal{L}_{reg}$ denotes the dual graph convolutional neural network loss function shown in (20), $\mathcal{L}_{GL}$ denotes the graph learning loss function shown in (26).

In summary, the flow of semi-supervised node classification algorithm based on GLDGCN is as follows

\begin{center}
\begin{algorithm}[ht]
\caption{Semi-supervised node classification based on Graph Learning Dual Graph Convolutional Neural Network (GLDGCN)} 
{\bf Input:} Adjacency Matrix $A \in \mathbb{R}^{n \times n}$ or Data Matrix $X = (x_1,x_2,...,x_n) \in \mathbb{R}^{n \times p}$. The label $Y = (x_1,x_2,...,x_n) \in \mathbb{R}^{n}$. Main Hyperparameters: Loss of PPMI convolutional $\lambda_1$ and loss of graph learning $\lambda_2$ \\
{\bf Output:} Classification Accuracy
\begin{algorithmic}[1]
\If{The input is adjacency Matrix $A \in \mathbb{R}^{n \times n}$}
\State $S_{ij} = h(x_i,x_j) = \frac{A_{ij}exp(ReLU(a^T|x_i - x_j|))}{\sum_{j=1}^n A_{ij}exp(ReLU(a^T|x_i - x_j|))}$ for the input of convolutional layers
\Else
\State $S_{ij} = h(x_i,x_j) = \frac{exp(ReLU(a^T|x_i - x_j|))}{\sum_{j=1}^n 
exp(ReLU(a^T|x_i - x_j|))}$ for the input of convolutional layers
\State \textbf{Return} The loss of graph learning layer $\mathcal{L}_{GL}$
\State $S$ for the GCN
\State \textbf{Return} The loss of GCN $\mathcal{L}_{0}$
\State $S$ for the PPMI convolutional layer
\State \textbf{Return} The loss $\mathcal{L}_{reg}$
\State mininize $\mathcal{L}_{0} + \lambda_1\mathcal{L}_{reg} + \lambda_2\mathcal{L}_{GL}$
\State \textbf{Return} The weights of GLDGCN
\State The neural network with trained parameters is used to predict the label of the prediction set, and compared with the real label, the classification Accuracy is obtained
\EndIf
\end{algorithmic}
\end{algorithm}
\end{center}

In GLDGCN, the graph convolutional neural network can be used for data in matrix form. The graph structure is mined by the graph learning layer, and then the convolutional layer and output layer are used for prediction. For data with graph structure itself, GLDGCN can also learn a better graph structure through the graph learning layer, which can be used as a more reasonable input to the convolutional layer. On the basis of classical message passing GCN, the dual convolution layer of GLDGCN further integrates the information of PPMI matrix, which is a more comprehensive extraction of graph structure information. In the next section, we will further demonstrate the advantages of GLDGCN through numerical experiments.

\subsection{Experiment}
To further demonstrate the effectiveness of the proposed Graph Learning Dual Graph Convolutional Neural Network (GLDGCN), we first conduct semi-supervised node classification experiments on the following three classical datasets.

\textbf{Citeseer}. This is an academic paper citation network with 3327 points and 4732 edges, which records the citation information between papers. There are 3703 dimensional feature vectors each node, indicating the existence/absence of corresponding words in the dictionary composed of 3703 unique words. The value of 0/1 is used as a word vector description, which is divided into 6 categories. Only 3.6\% of nodes have labels.

\textbf{Cora}. This is an academic paper citation network with 2708 points and 5429 edges, and there are 1433 dimensional feature vectors each node, which are divided into 7 categories. Only 5.2\% of nodes have labels.

\textbf{Pubmed}. This is an academic paper citation network with 19,717 points and 44,338 edges, and there are 500-dimensional feature vector each node, representing the frequency of occurrence in a dictionary of 500 terms, divided into 3 classes, with only 0.3\% of the nodes labeled for training.

\begin{table}[H]
\begin{minipage}{1.0\textwidth}
\centering
\caption{Three citation datasets}
\begin{tabular}{c|c|c|c|c|c}  
\toprule
Dataset   &  Nodes  & Edges & Classes & Features Dimension & Label Rate   \\ 
\toprule
Citeseer  & 3327 & 4732 & 6  & 3703 & 0.036                  \\
Cora      & 2708 & 5429 & 7  & 1433 & 0.052                 \\ 
Pubmed    & 19717 & 44338 & 3 & 500 & 0.003                    \\
\bottomrule 
\end{tabular}  
\end{minipage}
\end{table}

As can be seen from Table 1, the label rate of the three data sets is relatively low. There are few labeled samples, which requires higher semi-supervised classification capability of GCN. It should be noted that the setting of the training set follows the article \cite{kipf2016semi}\cite{yang2015network}. 20 topics in each category are selected from the three citation networks of Cora, Citeseer and Pubmed as the training set. In addition to the training set, 500 nodes were randomly selected as the verification set and 1000 nodes as the test set. The baseline methods compared also followed this setting. (Different technical divisions of the training/validation/test sets of the data set affect the prediction accuracy of the model\cite{zhu2020gssnn})

In our experiment, GLDGCN was built under the TensorFlow framework. The experimental software environment was Python 3.8(ubuntu20.04) and TensorFlow 1.15.5. The hardware environment is 64 vCPU AMD EPYC 9654 96-Core Processor, RTX 4090(24GB) * 4.

The main hyperparameters are set at dropout = 0.6, which randomly drops some neurons in the neural network during the training process, thereby reducing the complexity of the neural network, reducing the dependency between neurons, and preventing overfitting. Each layer has a weight attenuation coefficient of $5*10^{-3}$ and the training epoch of 1000. Dual convolutional neural network loss function coefficient $\lambda_1 = 0.01$, and graph learning layer loss function coefficient $\lambda_2 = 0.01$, the random walk step size is 3. The setting of other neural network architecture hyperparameters are adjusted with the change of dataset. The selection of architecture hyperparameters on different datasets is also discussed in the following through the analysis of experimental results.

\begin{table}[H]
\centering
\caption{The performance on three citation network(\%)}
\begin{tabular}{c|c|c|c}  
\toprule
Model   &                            Citeseer & Cora & Pubmed       \\ 
\toprule
DeepWalk\cite{perozzi2014deepwalk}  & 43.2 & 67.2 & 65.3            \\
Planetoid\cite{yang2015network}     & 64.7 & 75.7 & 77.2                  \\ 
Chebnet\cite{defferrard2016convolutional} & 69.8 & 81.2 & 74.4      \\
GCN\cite{kipf2016semi}             & 70.3 & 81.5 & 79.0                   \\
GAT\cite{velivckovic2017graph}     & 72.5 & 83.0 & 79.0    \\
LGCN\cite{gao2018large}              & 73.0 & 83.3  & 79.5             \\ 
GWNN\cite{xu2018graph}                  & 71.7 & 82.8  & 79.1            \\
GRNN\cite{ioannidis2019recurrent} & 70.8 & 82.8 & 79.5  \\
GOCN\cite{jiang2019robust}                 & 71.8    & 84.8  & 79.7       \\
GGD\cite{zheng2022rethinking}   & 73.0  & 83.9     & 80.5        \\
\bf{GLDGCN(ours)}                        & \bf{73.6} & \bf{85.8} & \bf{81.6}          \\
\bottomrule 
\end{tabular}  
\end{table}

It can be seen from GLDGCN designed in this paper for semi-supervised node classification has better performance than the baseline algorithm on the three citation datasets. Traditional random walk based algorithms, such as DeepWalk \cite{perozzi2014deepwalk} and Planetoid\cite{yang2015network}, in which DeepWalk cannot model attribute information. Planetoid has information loss in the process of random sampling. T.kipf \cite{kipf2016semi} uses convolution on undirected graphs to propose that classical message passing neural networks can greatly improve the classification accuracy of this task, but there are still defects that the information extracted by convolutional layers is relatively simple and must rely on graph data as input. The Graph Attention Network GAT\cite{velivckovic2017graph} can use the hidden self-attention layer to improve the classification accuracy by adjusting the weight of each feature in the training process. LGCN\cite{gao2018large} proposes the learning graph convolution layer LGCL automatically selects a fixed number of adjacent nodes for each feature according to the importance ranking, and the graph data is converted into a 1-D format grid-like structure, and regular convolution operations can be used on large graphs. However, the performance on the three classical data sets still has room for improvement. GWNN\cite{xu2018graph} improves the classification efficiency and interpretability of the network through the graph wavelet neural network, but at the same time, the generalization ability of the network is lost. GRNN\cite{ioannidis2019recurrent} enriches the input of feature matrices through propagation between layers, capturing multiple types of information. GOCN\cite{jiang2019robust} proposes a new potential graph convolution representation (LatGCR) for robust graph data representation and learning. The algorithm GLDGCN proposed in this paper combines the advantages of the above algorithms. It not only extends the input data that the network can receive from the graph to the general matrix form, but also extracts the input graph features comprehensively through the dual convolution layer. 

Then, we further discussed the influence of the number of training samples on the accuracy of classification results.

\begin{table}[H]
\centering
\caption{The influence of training sample points of Cora on the classification accuracy of GLDGCN (\%)}
\begin{tabular}{c|c|c|c|c|c|c|c|c|c}  
\toprule
The training samples(Cora)   &            27 & 54& 82 &        109   &  140    &    280     &   420    &   640    &  780        \\ 
\toprule
Classification Accurancy(\%)             &    67.9   &  78.3&79.9  &   82.8   & \textbf{85.8}       &  86.0  &  86.1    &     86.4   &   86.8  \\
\bottomrule 
\end{tabular}  
\end{table}

As can be seen from Table 3, taking Cora data set as an example, when the number of training samples accounts for 2\% or more of the total, GLDGCN proposed in this paper has good classification accuracy and strong model stability. This indicates that GLDGCN has low requirement for labeling rate and strong semi-supervised learning ability. It can also be seen from the changing trend of classification accuracy that the training sample proportion of 0.05 is also a reasonable choice, because when the number of training samples is too high, the classification accuracy of the model will be improved but not obvious, but it will also consume more computing resources.

We then compared the trend of classification accuracy of three models, GCN\cite{kipf2016semi}, GRNN\cite{ioannidis2019recurrent} and GLDGCN (this paper) with the proportion of training samples.
\begin{figure}[ht]
\centering
\includegraphics[width=0.6\textwidth]{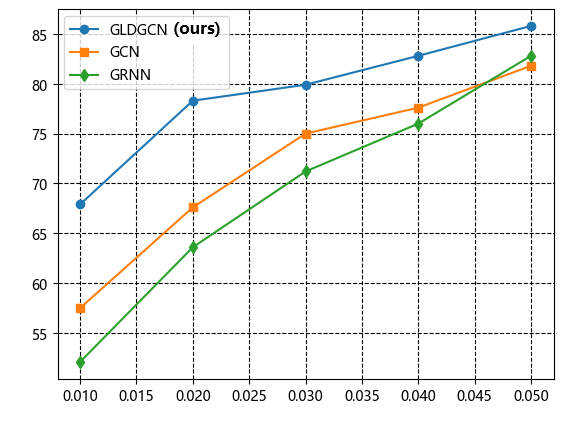}
\caption{The influence of training sample points of Cora on the classification accuracy of GLDGCN (\%)}
\end{figure}
In Figure 4, we take 1\%, 2\%, 3\%, 4\% and 5\% of the total training samples respectively on the Cora dataset to conduct experiments. The processing of verification set and test set remains unchanged, while the settings of each hyperparameter are not changed. It can be seen that the algorithm GLDGCN designed in this paper has a higher classification accuracy compared with other networks with a relatively small number of training tags, which indicates that the algorithm in this paper only needs to learn the relationship between node information through a very small number of tags, and can further reduce the cost of labeling in practical applications.

In addition, Zeng\cite{zeng2021decoupling} et al have pointed out that in large graphs, increasing the model depth of GCN usually means exponential expansion of the received information, which may not only cause the explosive growth of neighbor nodes, but also cause the performance of the network to decline due to excessive smoothing of node features. To this end, we also discussed the impact of the number of network layers on the classification accuracy of the model in this paper.

\begin{table}[H]
\centering
\caption{The influence of the number of convolutional layers on the classification accuracy of GLDGCN (\%)(OOM denotes Out Of Memory on 24GB RTX 4090)}
\begin{tabular}{c|c|c|c|c|c|c|c}  
\toprule
The number of layers   &                               2   &  3    &   4     &    5  &    6    & 7 &    8 \\ 
\toprule
Citeseer (\%)                         & 73.6 &  73.2  &  73.0   &   73.3  &   73.2   &  73.3   &   73.4    \\
Cora (\%)                             & 85.8 &  84.9  &  84.7   &   84.6  &   84.0   &    84.6 &  84.5     \\
Pubmed(\%)                           & 81.6 &  80.3  &  80.1   &   79.2  &   78.5   &    OOM &  OOM     \\   
\bottomrule 
\end{tabular}  
\end{table}

As can be seen from Table 4 of the experimental results, the classification accuracy of GLDGCN will decrease to a certain extent with the increase of the number of convolutional layers, but it is relatively stable. This also inspires us that, unlike classical convolutional neural networks, the increase of network layers in GCN may lead to decrease in learning performance. In fact, structural research on deep GCN is also a hot topic in recent years\cite{liu2020towards}. 

Among the three data sets, Cora(2708 points) and Citeseer (3327 points) are relatively small. The increase in the number of convolutional layers will decrease the prediction accuracy of the network, but the impact is not significant. Pubmed(19717 points) is relatively large. For large graph, GLDGCN runs slowly and requires large computing resources, and the problem of insufficient GPU memory often occurs in the training process. When the number of convolutional layers is increased to more than 7 layers, GPU has a phenomenon of insufficient memory, which indicates that when the network performs experiments on large graphs, attention should be paid to control the number of convolutional layers to prevent the problem of excessive algorithm space complexity caused by the explosive growth of neighbor nodes. It is basically consistent with Zeng \cite{zeng2021decoupling}. 

In addition, several experiments were carried out on Cora and Citeseer data sets, and the relationship between the value of each hyperparameter and classification accuracy was statistically analyzed, aiming to provide some insights for the selection of hyperparameters of the model. Due to the large Pubmed data set, slow model running speed, and the problem of insufficient video memory likely to occur during operation, the results of multiple experiments may not be stable enough, so the analysis is not carried out in this part.

\begin{figure}[ht]
\centering
\includegraphics[width=0.75\textwidth]{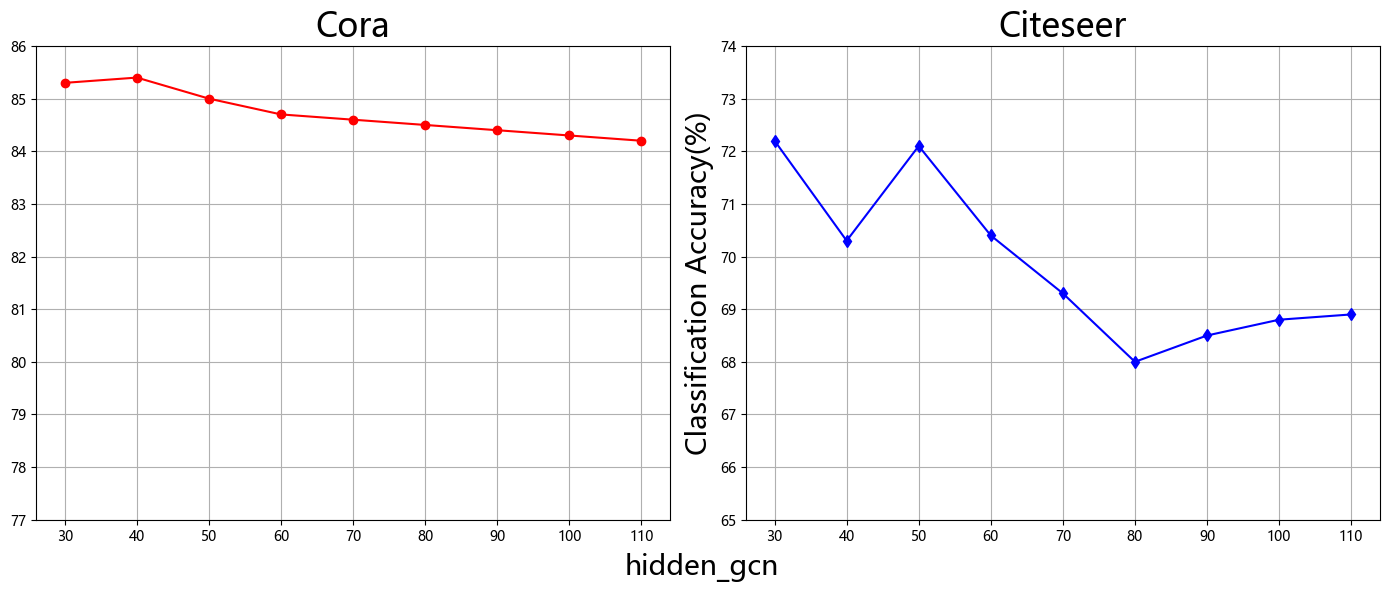}
\caption{The influence the hyperparameters hidden\_gcn on the classification accuracy of GLDGCN (\%)}
\end{figure}

As can be seen from Figure 5, the classification prediction accuracy of GLDGCN is generally stable on Cora, and is basically not affected by the change of the number of hyperparameter hidden layer elements hidden\_gcn. For Citeseer, a reasonable value such as 30 or 50 should be set for this hyperparameter to achieve a better prediction effect.

\begin{figure}[ht]
\centering
\includegraphics[width=0.75\textwidth]{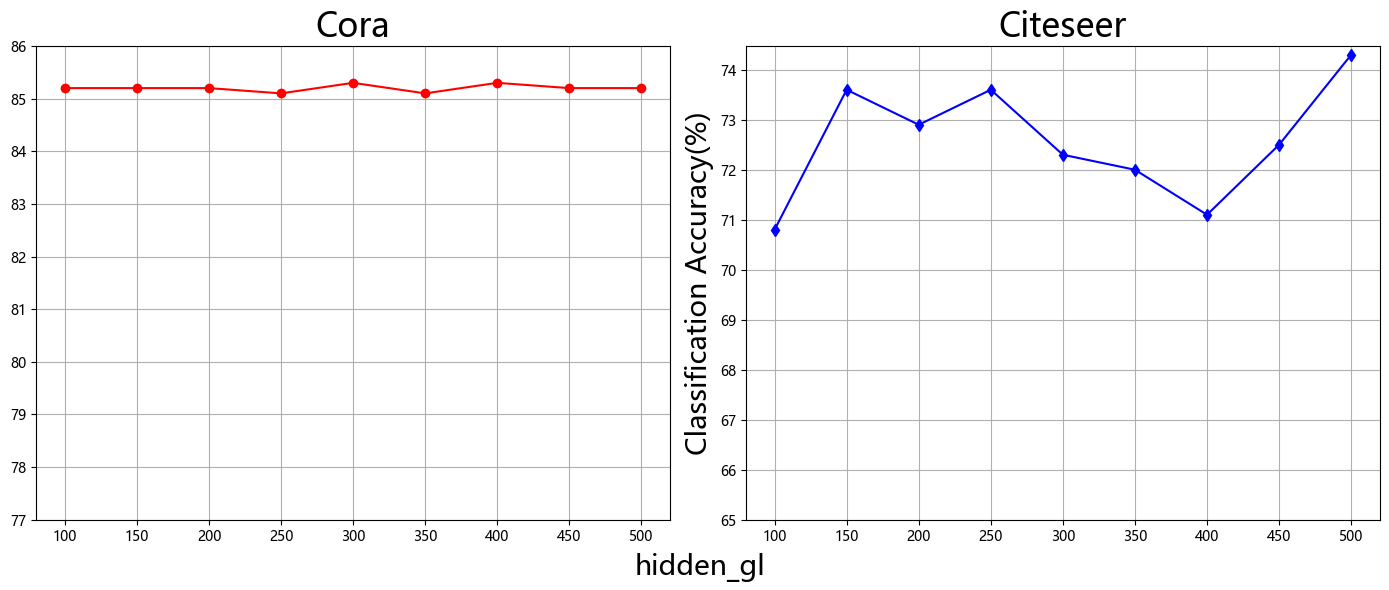}
\caption{The influence the hyperparameters hidden\_gl on the classification accuracy of GLDGCN (\%)}
\end{figure}
Similar to Figure 5, in Figure 6, the classification prediction accuracy of GLDGCN is generally stable on Cora, and is basically not affected by the change of the number of hyperparameter hidden layer elements hidden\_gl. For the Citeseer, attention should be paid to setting the value of this hyperparameter to obtain better prediction accuracy. In addition, during the experiment, we found that when the number of hidden layer elements hidden\_gl is greater than 400, the model becomes non-convergence on Citeseer, so it is recommended to set hidden\_gl between the interval [150,250].
\begin{figure}[ht]
\centering
\includegraphics[width=0.75\textwidth]{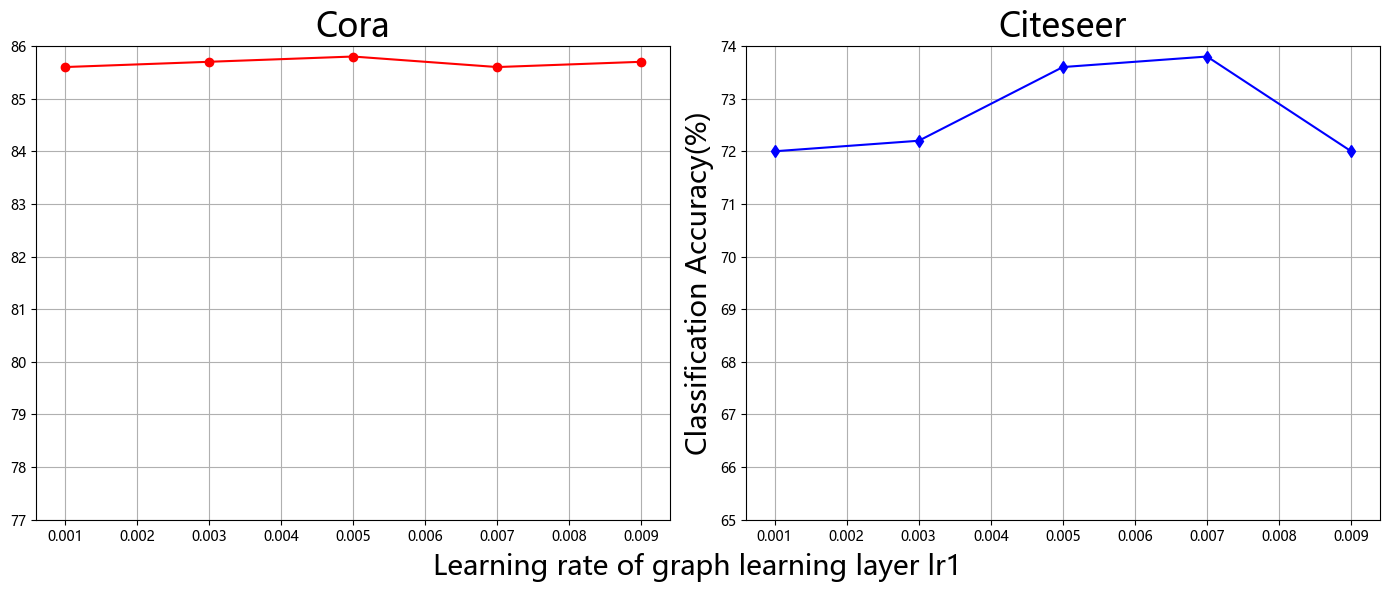}
\caption{The influence the hyperparameters lr1 on the classification accuracy of GLDGCN (\%)}
\end{figure}
\begin{figure}[ht]
\centering
\includegraphics[width=0.75\textwidth]{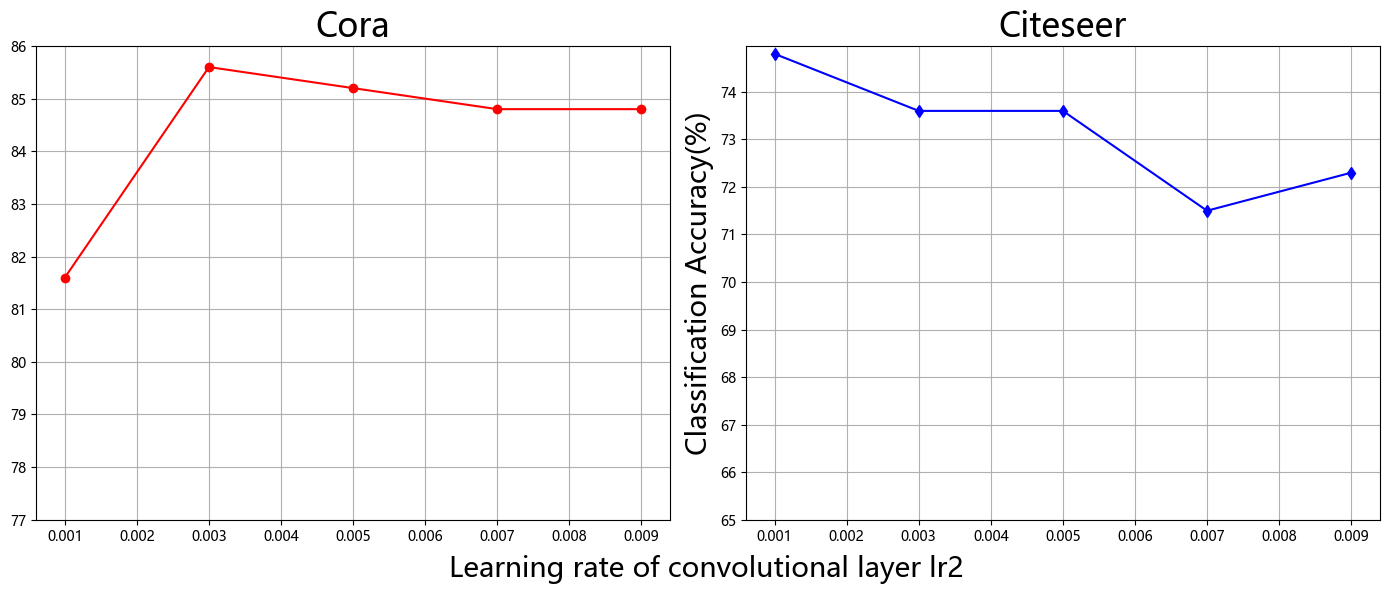}
\caption{The influence the hyperparameters lr2 on the classification accuracy of GLDGCN (\%)}
\end{figure}

Figure 7 shows that on Cora data set and Citeseer data set, the classification prediction accuracy of GLDGCN is generally stable, and it is basically not affected by the learning rate lr1 of the hyperparameter graph learning layer, indicating that the model has strong stability on this hyperparameter. Selecting lr1 in [0.005,0.007] gives the model better classification performance on the Citeseer.

Figure 8 analyzes the influence of hyperparameter convolutional layer learning rate lr2 on model prediction accuracy. For Cora, if convolutional layer learning rate lr2 is selected in the interval [0.003,0.007], the model prediction accuracy will be relatively stable, which is conducive to the model's good prediction performance. For Citeseer, lr2 can be set to 0.001 and below to pursue higher classification accuracy.

In general, graph learning layer learning rate lr1 and convolutional layer learning rate lr2 are two parameters that have relatively little impact on model classification accuracy. For different datasets, they only need to be set within a reasonable interval.

In order to further explore the generalization ability of the algorithm, GLDGCN is also used on the classical small dataset KarateClub and the large dataset WikiCS. 

\textbf{KarateClub}. The classic Zachary karate club network \cite{zachary1977information}, contains 34 nodes, connected by 156 (undirected and undirected) edges each node is labeled by one of four classes based on modular clustering. The training is based on a single tag example for each class, that is, a total of four tag nodes. GLDGCN only need at least 1 label per class, \textbf{to classify all nodes correctly.}

\textbf{Wiki-CS}\cite{mernyei2020wiki}. Semi-supervised classification data set based on Wikipedia, containing 11,701 nodes, 216,123 edges, divided into 10 classes, each node with 300 dimensional feature, node labeling rate of 5\%. Figure 9 shows a visual comparison between Wiki-CS and the three classic reference networks. It can be seen that the number of nodes and edges of Wiki-CS are large, and the connections between nodes are complex, making classification more difficult than that of the three classic reference networks.

\begin{figure}[ht]
\centering
\includegraphics[width=0.75\textwidth]{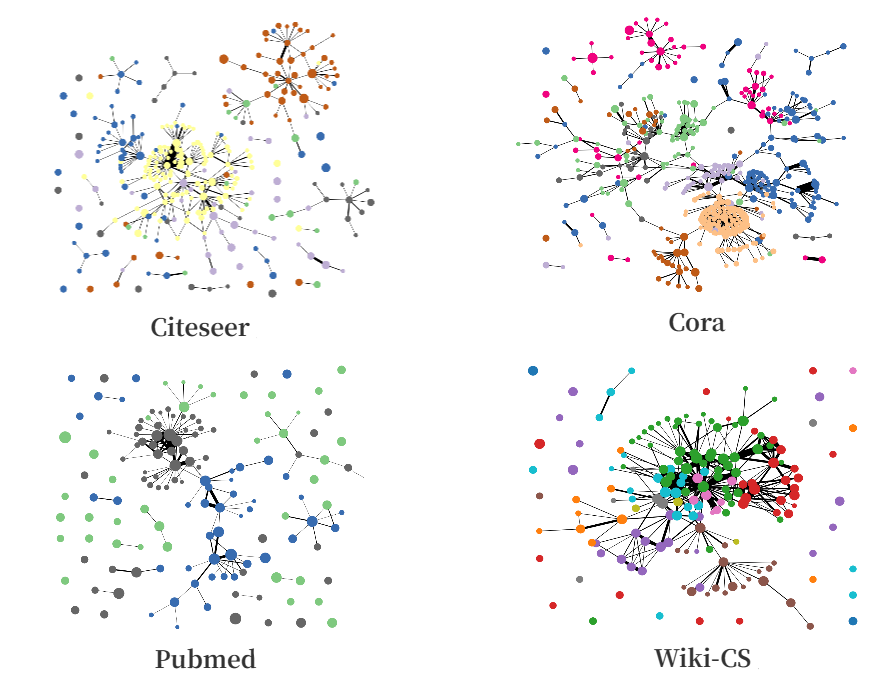}
\caption{A visual comparison of WiKi-CS with three classical reference networks \cite{mernyei2020wiki}, where each node corresponds to a cluster of similar nodes in the original image, edge thickness represents the amount of connections between the clusters, and color represents different classifications}
\end{figure}

As can be seen from Table 5, the proposed algorithm (GLDGCN) still perform better compared with the baseline algorithm that completes the node classification task on the WiKi-CS dataset. Although only a small part of the node information (5\%) of the data set is used in the training process, GLDGCN can still effectively learn the internal connections between node features from the complex node connections and enhance the classification ability.

\begin{table}[ht]
\centering
\caption{Semi-supervised node classification accuracy(\%) on Wiki-CS}
\begin{tabular}{c|c}  
\toprule
Model   &                                          Wiki-CS  (\%)   \\ 
\toprule
SVM\cite{jakkula2006tutorial}                         & 72.6      \\
MLP\cite{taud2018multilayer}                 & 73.2              \\
GCN\cite{kipf2016semi}                         & 79.1           \\
GAT\cite{velivckovic2017graph}                 & 79.7          \\
APPNP\cite{klicpera2018personalized}             & 79.8         \\
\bf{GLDGCN(ours)}                             & \bf{81.2}         \\
\bottomrule 
\end{tabular}  
\end{table}

\section{TRAINING ON LARGE GRAPHS}
Since graph convolution operators in GCN need to propagate features through information interaction between the nodes in the graph, loss terms in GCN (for example, classifying the loss of a single node) depend on a large number of neighbouring nodes. Because of this dependency, the backpropagation process requires that all embeddings in the computed graph be stored in GPU memory during training, which leads to low efficiency.

Taking the input as $X = (x_1,x_2,...,x_n) \in \mathbb{R}^{n \times p}$, the number of nodes in the figure is $N$, the node feature is $p$ dimension, and $L$ represents the number of convolutional layers in GCN. The training method of full batch gradient descent in \cite{kipf2016semi} is adopted. The space complexity required for the computation is $\mathcal{O}(NpL)$, which puts a high demand on the GPU memory.

The GLDGCN in this paper also has such a problem, when the data set is a large graph, for example, with more than 20,000 nodes, the RTX 4090 * 4 with 24 GB of memory is barely enough to meet the storage requirements. But in the real world, large graphs are everywhere, so
we design an algorithm to train graph convolutional neural networks using subgraph clustering techniques. The aim of this algorithm is to construct node partitions, which are subgraphs of nodes in different regions for parameter updating during training. The algorithm is based on the efficient graph clustering algorithm METIS\cite{karypis1998fast} to design batches, combined with stochastic gradient descent (SGD) techniques to reduce the spatial overhead of traditional graph convolutional neural networks.

This algorithm can bring relatively large memory and computing advantages. In terms of memory, only the node in the current mini-batch need to be stored, the space complexity is $\mathcal{O}(bpL)$, and $b \ll N$ is the batch size, which is superior to other training methods based on full gradient descent. In terms of computational complexity, with the same time cost, the algorithm can update the parameters with less space complexity through stochastic gradient descent. This makes GLDGCN designed in this paper applicable to larger graph datasets such as PPI dataset and Reddit dataset.

\subsection{Clustering-GCN}
For the undirected graph $\mathcal{G} = (\mathcal{V}, \mathcal{E})$, we divide the nodes into c groups $\mathcal{V} = [\mathcal{V}_1, ... , \mathcal{V}_c]$, where $\mathcal{V}_t$ denotes the $t$-th node set. We can consider a graph $\mathcal{G}$ consisting of its c subgraphs
\begin{equation}
\bar{G} = [G_1,...G_c] = [\{\mathcal{V}_1,\mathcal{E}_1\},...,\{\mathcal{V}_c,\mathcal{E}_c\}],
\end{equation}
Each $\mathcal{E}_t$ contains only the edges that exist in the nodes of $\mathcal{V}_t$. After dividing the nodes, the adjacency matrix $\bf{A}$ of the original graph is also divided into $c^2$ submatrices
\begin{equation}
A = \bar{A} + \triangle = 
\begin{bmatrix}
A_{11}  & \dots & A_{1c} \\
\vdots &  \ddots & \vdots \\
A_{c1} & \dots & A_{cc} \\
\end{bmatrix},
\end{equation}
and
\begin{equation}
\bar{A} = \begin{bmatrix}
A_{11}  & \dots & 0 \\
\vdots &  \ddots & \vdots \\
0 & \dots & A_{cc} \\
\end{bmatrix},
\triangle = 
\begin{bmatrix}
0  & \dots & A_{1c} \\
\vdots &  \ddots & \vdots \\
A_{c1} & \dots & 0 \\
\end{bmatrix}.
\end{equation}

Each diagonal element $A_{tt}$ is a diagonal matrix of $|\mathcal{V}_t| \times |\mathcal{V}_t|$, containing all the edges in $\hat{G}_t$. $\bar{A}$ is the adjacency matrix of the graph $\bar{G}$. $A_ {st}$ contains the child point set $\mathcal{V}_s$ and $\mathcal{V}_t$ formed by the edges. $\triangle$ is a matrix of all the non-diagonal elements in $A$. Similarly, We can divide feature matrix $X$ and labels $Y$ according to $\mathcal{V} = [\mathcal{V}_1, ... , \mathcal{V}_c]$ into  $[X_1, ... , X_c]$ and $[Y_1,... Y_c]$, where $X_t$ and $Y_t$ contain the features and labels of all the nodes in $\mathcal{V}_t$.

The advantage of this division is that the forward propagation process of the graph convolutional neural network can also be divided into different batches.
\begin{equation}
\begin{split}
   Z  = \bar{A}^{'}\sigma(\bar{A}^{'}\sigma(\dots \sigma(\bar{A}^{'}XW^{(0)})W^{(1)})\dots)W^{(L-1)} \\
      = \begin{bmatrix}
\bar{A}_{11}^{'}\sigma(\bar{A}_{11}^{'}\sigma(\dots \sigma(\bar{A}_{11}^{'}XW^{(0)})W^{(1)})\dots)W^{(L-1)} \\\\
\cdots  \\\\
\bar{A}_{cc}^{'}\sigma(\bar{A}_{cc}^{'}\sigma(\dots \sigma(\bar{A}_{cc}^{'}XW^{(0)})W^{(1)})\dots)W^{(L-1)} \\
\end{bmatrix}.
\end{split}
\end{equation}
$\bar{A}^{'}$ denotes the normalization of $\bar{A}$ with $\bar{A}^{'}$ = $\bar{D}^{-\frac{1}{2}} \bar{A}\bar{D}^{-\frac{1}{2}}$.
The loss function also change into 
\begin{equation}
\mathcal{L}_{\bar{A}^{'}} =  \sum_{t} \frac{|\mathcal{V}_t|}{N} \mathcal{L}_{\bar{A}^{'}_{tt}}
\quad
and
\quad
\mathcal{L}_{\bar{A}^{'}_{tt}} =  -\sum_{l \in \mathcal{V}_t} \sum_{f=1}^{d_{out}} \mathcal{Y}_{l,f}ln Z_{l,f}.
\end{equation}

\begin{figure}[H]
\centering
\includegraphics[width = 0.75\textwidth]{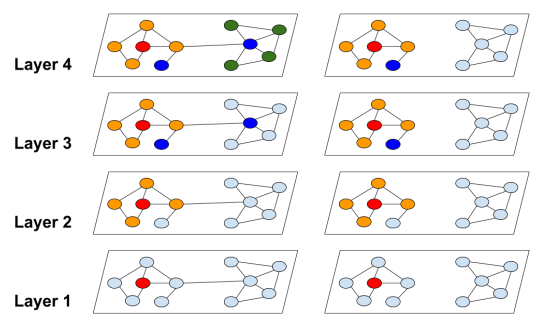}
\caption{The left figure is the node propagation mode of traditional GCN. When facing large size graphs or using deeper structure, the problem of neighbor node expansion will occur, resulting in high GPU memory requirements. The figure on the right shows that after the introduction of subgraph clustering technology, the network can avoid dependence on neighbor nodes between subgraphs during training and reduce the consumption of computing resources\cite{chiang2019cluster}}
\end{figure}

\subsection{GCN with Subgraph Coarsening}
The introduction of clustering technology also enables us to design personalized cluster graph convolutional neural networks according to the characteristics of different subgraphs in different data sets, so as to improve the generalization ability of GCN when performing node classification tasks on specific data sets, which is similar to the idea of designing graph convolutional neural networks based on the structure of subgraphs.

Figure 11 shows a general framework for cluster graph convolutional neural networks. When the same GCN framework is used for every cluster, such as messaging GCN or GLDGCN as presented in  this paper, GCN has the ability to perform semi-supervised learning on large graphs. But there are the following problems.

1. If nodes with the same label are clustered together in the clustering process, the features of each small-batch training set may be inconsistent with the features of the whole graph, resulting in Biased SGD \cite{demidovich2024guide}.

2. Since large graphs are input into GCN in the form of subgraphs, in order to avoid neighbor node explosion caused by training deep GCN, nodes between different subgraphs cannot achieve message transfer, which will greatly weaken the ability of GCN to extract long-distance (nodes between different subgraphs) information, which may affect the generalization ability of GCN.

\begin{figure}[ht]
\centering
\includegraphics[width=0.75\textwidth]{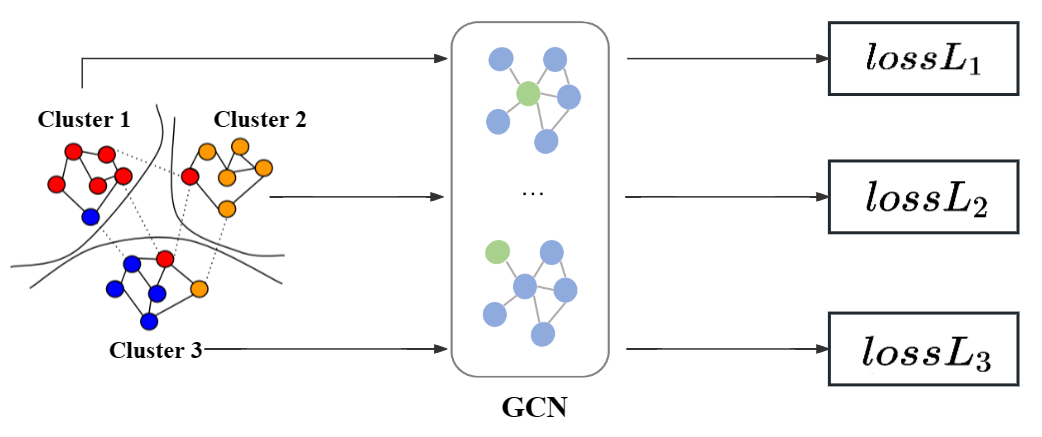}
\caption{A general framework for semi-supervised node classification based on cluster-gcn}
\end{figure}

Therefore, as shown in our algorithm, we start by dividing the input large graph into $c$ subgraphs. For each $q$ subgraph, the GLDGCN is shown in Figure 3, and then the parameters are updated by minimizing the loss function. According to the stochastic gradient descent rule, all subgraphs can be trained with sufficient parameters. After the molecular map is delimited by METIS algorithm, $q$ classes can be randomly selected to form subgraphs as a batch of training, which ensures that the types of labels in each batch are uniform enough to a certain extent. 

\begin{center}
\begin{algorithm}[ht]
\caption{Semi-Supervised Node Classification Algorithm on Large Graphs Based on Clustering Graph Convolutional Neural Networks} 
\textbf{Input:} Large Graph $\mathcal{G}$, Feature matrix $X$, Label $Y$\\
\textbf{Output:} Classification Accuracy 
\setstretch{1.0} 
\begin{algorithmic}[1]
\State Using METIS\cite{karypis1998fast} to divide the graph into $\mathcal{V}_1, ... , \mathcal{V}_c$
\For{Each node \( x_i \in \mathcal{V} \)}
    \State From \( x_i \) 
    \For{$iter = 1, \dots , max_{iter}$}
        \State Select q classs $t_1, ... , t_q$, $q \ll c$ \text{from} $\mathcal{V}$
        \State Subgraphs with node Set $\bar{\mathcal{V}} = [\mathcal{V}_{t_1}, \mathcal{V}_{t_2} ... , \mathcal{V}_{t_q}]$ and adjacency matrix $A_{{\bar{\mathcal{V}}},\bar{\mathcal{V}}}$
    \For{each $A_{{\bar{\mathcal{V}}},\bar{\mathcal{V}}}$ of subgraph }
    \State Computing $S_{ij} = h(x_i,x_j) = \frac{A_{ij}exp(ReLU(a^T|x_i - x_j|))}{\sum_{j=1}^n A_{ij}exp(ReLU(a^T|x_i - x_j|))}$ as the input of the convolutional layer
    \State \textbf{Return} The loss function of graph learning layer $\mathcal{L}_{GL}$
    \For{each $S_{\bar{\mathcal{V}},\bar{\mathcal{V}}}$}
        \State input $S_{\bar{\mathcal{V}},\bar{\mathcal{V}}}$ to GCN according to (27)
        \State \textbf{Return} The loss function of GCN $\mathcal{L}_{0}$ 
        \State input $S_{\bar{\mathcal{V}},\bar{\mathcal{V}}}$ to PPMI convolutional layer according to (28)
        \State \textbf{Return} The loss function of PPMI Convolutional Layer $\mathcal{L}_{reg}$ 
        \State minimize $\mathcal{L}_{0}$ + $\lambda_1\mathcal{L}_{reg}$ + $\lambda_2 \mathcal{L}_{GL}$
        \State \textbf{Return} The weights of neural network
     \EndFor
   \EndFor 
 \EndFor
\EndFor
        \State Training stops until all subgraphs are used for parameter update and the changes do not exceed a certain threshold
        \State GCN with trained parameters is used to predict the prediction set to output semi-supervised classification accuracy 
\end{algorithmic}
\end{algorithm}
\end{center}

Huang et al\cite{huang2021scaling} and Liu et al\cite{Liu2024ICLR} introduce the graph coarsening technique. This technology can be combined with subgraph clustering technology to make GCN extract long-range information, so as to make up for the defects of mini-batch processing.

\begin{figure}[H]
\centering
\includegraphics[width=0.75\textwidth]{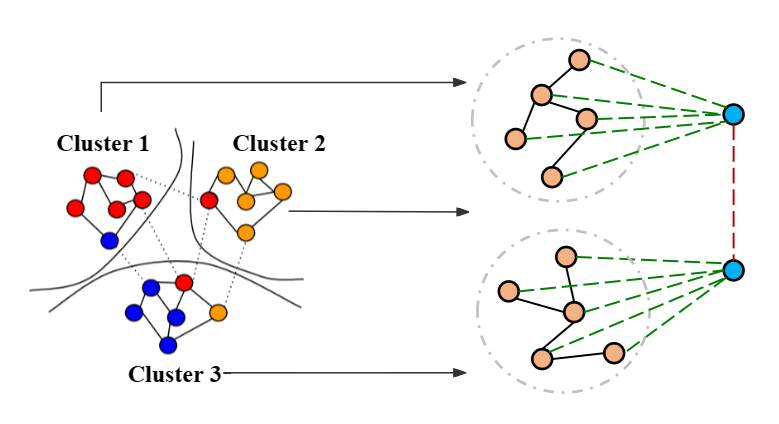}
\caption{Each class can be regarded as a subgraph in a large graph, and then each subgraph is roughed up separately, adding coarse nodes to represent subgraph information, edges between coarse nodes and coarse nodes represent information between subgraphs, edges between coarse nodes and subgraph nodes record information of the graph coarsing process, and are used to generate coarse node embeddings from subgraph nodes}
\end{figure}

\subsection{Experiment}
The introduction of subgraph clustering technology enables us to train deeper GCN on Pubmed (19717 nodes). 

\begin{table}[ht]
\centering
\caption{The effect of the number of convolutional layers on the classification accuracy of Pubmed (\%) }
\begin{tabular}{c|c|c|c|c|c|c|c|c|c}  
\toprule
The numbers of convolutional layers   &    2   &  3 & 4  &  5  &  6  &  7  &  8  & 9 & 10 \\ 
\toprule
GCN                      & 79.0 &  78.8  &  78.7   &   78.5  &   79.0  & 78.7  & 78.5 &  78.7 &  78.9 \\
\bf{CLGCN(2 batches)}            & 79.3 &  79.0  &  78.8   &   79.2  &   78.9   &  79.0 &  79.1 & 78.9 &  79.1   \\   
\bf{CLGCN(4 batches)}            & 78.9 &  78.8  &  79.2   &   79.0  &   79.1   &  78.9  &  78.8 & 78.9 &  79.3   \\  
\bf{CLGCN(8 batches)}            & 79.2 &  79.3  &  79.1   &   79.1  &   79.0   &  78.8  &  79.1 & 79.0 &  79.5   \\  
\bottomrule 
\end{tabular}  
\end{table}
In Table 4 of this paper, when the algorithm GLDGCN is used to train neural networks with more than 7 layers on Pubmed, the problem of insufficient GPU memory (OOM) appears. As can be seen from Table 6, after the introduction of subgraphs clustering technology, we have been able to design 10 layers or even deeper GCN. Not only does OOM phenomenon not occur, but also the classification accuracy has been improved compared with the classical full-batch training method.

The introduction of subgraph clustering technology realizes the mini-batch training of traditional GCN model, which may lose the effectiveness of the original GCN intuitively. The reason is that these methods divide the entire graph into several subgraphs as small batches, which prevents information from propagating between different small batches. However, it can be seen from Table 6 that with the increase of the number of network layers, small-batch training technology will improve the classification accuracy of GCN. This is due to the fact that after mini-batch processing, the model better learns the subgraph structure features of the raw data \cite{fey2021gnnautoscale}. The more mini-batches, the easier it is for GCN to learn important subgraph structures in the dataset, improving its node classification capabilities. The recognition of subgraph structure is very important to improve the expressive ability of GCN, which can help it solve the graph isomorphism problem \cite{zhang2023complete} more effectively in small datasets. Due to the large Pubmed dataset and the main node classification tasks, the improvement of the generalization ability is not obvious, but the relationship between the representation ability in large graphs and its generalization ability of GCN in node classification tasks is still an Open Question. The introduction of subgraph clustering technology can not only help train deeper GCN, enhancing the ability of GCN to complete node classification tasks on large graphs, but also help GCN to identify more key subgraph structures, improving the expressive ability of GCN, and thus enhance the interpretability of GCN.

In order to further explore the ability of clustering GCN to process large graphs, we also conducts corresponding experiments on larger PPI data sets and Reddit data sets.

\textbf{PPI} is a protein-protein interaction dataset, expressed in graph form, where nodes represent proteins and edges represent interactions between proteins. These data sets can be real data from laboratory measurements or predictive data based on calculations. \textbf{There are 56,944 points, 818,716 edges, each node has 50 dimensional features, and is divided into 121 classes}. The experimental data show that in the process of graph clustering, PPI data set is divided into 50 subgraphs, and one subgraph is taken each time for training and updating parameters.

\textbf{Reddit}. Reddit is a well-known social news aggregator, community and discussion site commonly used for tasks such as social network analysis, natural language processing, sentiment analysis, topic modeling, and more. These data sets can contain a variety of different information, such as the text content of the post, when it was posted, author information, comment content, voting status, and so on. \textbf{There are 232,965 points, 1160,6916 edges, each node has 602 dimensional features, and is divided into 41 categories}. The experimental data shows that in the process of graph clustering, the Reddit data set is divided into 1500 subgraphs, and 20 subgraphs are taken each time for training and parameter updating.

\begin{table}[H]
\centering
\caption{Semi-supervised Node Classification Accuracy on Large Graph PPI and Reddit 
(OOM denotes insufficient GPU memory on RTX 4090 with 24GB )}
\begin{tabular}{c|c|c}  
\toprule
Model   &                                PPI (\%) & Reddit (\%)        \\ 
\toprule
GCN\cite{kipf2016semi}          & OOM &  OOM               \\
GraphSAGE\cite{hamilton2017inductive}   & 61.2 & 95.4    \\
GAT\cite{velivckovic2017graph}    & 97.3 & OOM            \\
VR-GCN\cite{chen2017stochastic}        &  97.0  &  96.1      \\
GraphSAINT\cite{zeng2019graphsaint}     &  98.1   &  96.5          \\
VQ-GNN\cite{ding2021vq}                & 97.4  &  94.5     \\
\bf{CLGCN(ours)}                       & \bf{98.6} & 96.2             \\
\bottomrule 
\end{tabular}  
\end{table}

As can be seen from Table 7, GCN designed based on subgraph clustering technology can still perform well on large graph. We \textbf{rank first in PPI data set and second in Reddit data set}. The introduction of subgraph clustering technology also enables GCN to process large-scale graph data, expands the application value of GCN.

\section{CONCLUSION}
Setting out from the most classical GCN, we design a graph convolutional neural network GLDGCN by introducing dual convolutional layer and graph learning layer, and apply it to the classical semi-supervised node classification task, and designs a graph convolutional neural network algorithm for semi-supervised node classification. In addition, in order to solve the problem that the algorithm has insufficient ability to process large graphs during the experiment, we also introduce subgraph clustering technology into GCN, and propose a semi-supervised node classification algorithm based on clustering GCN, which makes the network have the ability to process large graphs.

However, this paper also has a lot of room for optimization. For example, more theoretical analysis can be carried out in the aspects of algorithm hyperparameter selection and convergence, so as to further provide the interpretability of the algorithm; In network design, a more personalized GCN can be designed for different data sets, and a deeper convolutional neural network can be designed to adapt to more complex feature processing tasks. The experiment in this paper has some problems such as too small sample size of some data sets and insufficient rationality of feature engineering, which will reduce the necessity and scientificity of algorithm design.

\bibliographystyle{unsrt}  
\bibliography{references}  

\end{document}